\documentclass[conference]{IEEEtran}
\usepackage{times}

\usepackage[numbers]{natbib}

\usepackage{amsmath,amssymb,amsfonts}
\usepackage{amsthm}

\usepackage{graphicx}
\usepackage{booktabs}
\usepackage{xcolor}
\usepackage{microtype}
\usepackage{tikz}
\usepackage{fontawesome5}
\usetikzlibrary{arrows.meta, positioning, fit, backgrounds, calc}

\usepackage[bookmarks=true,colorlinks=true,allcolors=blue]{hyperref}
\usepackage[capitalize]{cleveref}

\newcommand{\xt}{\mathbf{x}_t}
\newcommand{\xz}{\mathbf{x}_0}
\newcommand{\xone}{\mathbf{x}_1}
\newcommand{\xhat}{\hat{\mathbf{x}}_0}
\newcommand{\vfield}{\mathbf{v}}
\newcommand{\grad}{\nabla}

\newcommand{\R}{\mathbb{R}}

\begin{document}

\title{Is Energy Guidance All You Need?\\
Training-Free Norm Injection for Driving World Models}

\author{Xiyan Su, Frank Diermeyer, Markus Lienkamp \\
Institute of Automotive Technology, Technical University of Munich, Munich Germany}

\maketitle

\begin{abstract}
Driving world models built on large video-diffusion backbones generate realistic scenes but are hard to control: enforcing a traffic norm typically means retraining the backbone or conditioning it on hand-built layouts. We ask whether controllability requires training at all. Our experiment shows that a rectified-flow driving world model, which jointly generates future video and a planned ego trajectory, can have its planned trajectory steered entirely at sampling time by differentiable energy functions that encode driving norms, without knowledge-specific retraining of the diffusion backbone. Concretely, we demonstrate that a world model built on Open-Sora~2.0 MM-DiT backbone can be steered to brake at a counterfactual target by injecting energy guidance at sampling time. However, we find that the generated video does not yet follow the steered trajectory through the backbone's joint self-attention and identify the cross-stream coupling as a crucial requirement for end-to-end-controllable rollouts.
\end{abstract}

\IEEEpeerreviewmaketitle

\section{Introduction}
\label{sec:intro}

Video-diffusion driving world models generate realistic sensor streams and have become increasingly popular for simulation and planning \citep{hu2023gaia1,gao2024vista,wang2024drivewm}. Their central practical weakness is controllability: a model that produces a beautiful, crisp scene in which the ego vehicle drifts across a solid line, exceeds the speed limit, or collides with another agent is not useful for safety-critical use. However, state-of-the-art mitigation strategies remain costly to scale. One either fine-tunes the backbone whenever a new constraint must be complied with, or one conditions the model on hand-constructed layouts baked in during training, such as maps and agent bounding boxes. Both require high-quality data and couple new driving knowledge to a training run. Neither lets a deployed model add or re-prioritize a rule during sampling time.

\begin{figure}[t]
\centering
\includegraphics[width=\columnwidth]{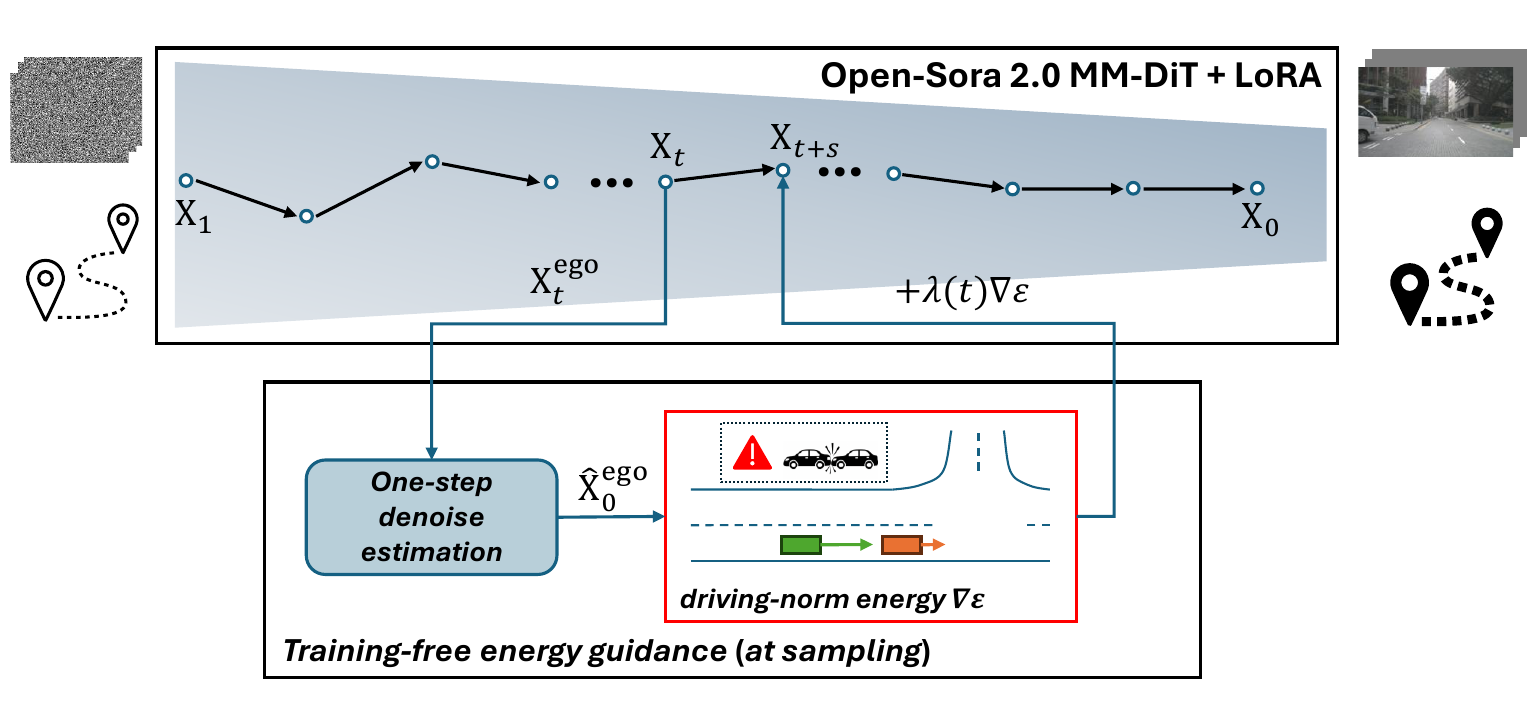}
\caption{\textbf{Overview.} A frozen video world model (Open-Sora~2.0
MM-DiT\,+\,LoRA) jointly denoises future video and an ego trajectory conditioned only on observed past without agent or map at training. Driving norms enter at sampling
as energies on the clean ego estimate $\xhat$ and are applied as a
velocity correction to the ego stream. We study the question of whether the generated video is also steered through the backbone's joint self-attention.}
\label{fig:overview}
\end{figure}

\textbf{Hypothesis:} We ask whether controllability requires training at all. Training-free guidance has shown that a frozen diffusion prior can be steered toward samples that satisfy an external objective by perturbing the sampling trajectory with the gradient of a differentiable criterion \citep{yu2023freedom,bansal2023universal,chung2023dps,feng2025guidanceflowmatching}. We bring this view to world models for autonomous systems: as an application example, we take driving norms as energy functions evaluated on the generated ego trajectory, and inject them only at sampling time into a frozen rectified-flow backbone. The backbone is trained completely without knowledge-specific conditioning (see \cref{fig:overview}). 

\textbf{Does video match trajectory?} A driving world model is valuable precisely because it generates the video, not just a trajectory. Therefore, the question we center on is whether steering the ego trajectory with an energy term also guides the video to match through the backbone's joint denoising. We find that the trajectory is steered readily, but the video does not yet follow. We trace this back to the model architecture and identify the cross-stream coupling as the main cause.

\textbf{Contributions:}
Our contributions are three-fold:
\begin{itemize}\itemsep2pt
  \item We frame driving-norm compliance as training-free energy guidance on a frozen rectified-flow world model that jointly generates future video and a planned ego trajectory, and we show that all the knowledge can be added only at the sampling time and how.
  \item We test our hypothesis on a world model built on an Open-Sora~2.0 backbone \citep{peng2025opensora2} and show that a training-free energy guidance steers the ego trajectory to a counterfactual braking target at sampling time, simulating a rear-end collision driving norm.
  \item We identify the central obstacle to end-to-end control: the steered ego trajectory does not propagate to the generated video under the current model architecture, i.e., double-stream rectified flow model aligned through joint self-attention. Based on this finding, we propose a possible solution to that problem.
\end{itemize}

\section{Related Work}
\label{sec:related}

\textbf{Training-free diffusion guidance.} A frozen diffusion prior can be steered toward an external objective by perturbing the sampling trajectory with the gradient of a differentiable criterion evaluated on the predicted clean sample $\xhat$~\citep{yu2023freedom,bansal2023universal,chung2023dps,ye2024tfg}. Energy- and composition-based variants combine several criteria \citep{liu2022composable,du2023reduce}. These methods are known to drift off the data manifold and to suffer when $\xhat$ averages distinct modes \citep{chung2023dps,he2024mpgd}. We apply a similar method to steer the ego plan of a frozen video world model rather than a low-dimensional trajectory model.

\textbf{Guidance for flow matching.} We use a rectified-flow model as our backbone~\citep{liu2023rectifiedflow,lipman2023flowmatching,albergo2023stochastic}, whose deterministic ODE velocity field requires a different guidance treatment than score-based SDEs. Feng et al. \cite{feng2025guidanceflowmatching} formalize guidance for flow matching. We build on this and apply it to a world model. Concurrent work couples flow matching with a trained energy model for ego trajectory planning~\citep{liu2025guideflow}. We instead inject energies training-free at sampling time without retraining.

\textbf{Controllable video diffusion.} Training-free and conditional control of video diffusion has targeted motion, trajectories, and frame appearance \citep{jang2025frameguidance,wang2024motionctrl,yin2023dragnuwa,wang2024boximator}. These control how the video looks or moves. We instead inject semantic driving norms into a video world model.

\textbf{Robotic World Models for Driving.} Video-generation models produce realistic rollouts and, increasingly, actions for autonomous driving and general robotics \citep{hu2023gaia1,gao2024vista,wang2024drivewm,russell2025gaia2,wang2024drivedreamer}. In this field, control is often achieved by training-time structured conditioning (maps, layouts; e.g.\ GAIA-2 \citep{russell2025gaia2}) or by generate-then-select over sampled futures (e.g.\ Drive-WM \citep{wang2024drivewm}). Both approaches require additional cost. Recent models jointly generate the ego trajectory alongside the video~\citep{zhang2025epona,xia2025drivelaw}, one on a rectified-flow backbone like ours~\citep{xia2025drivelaw}. Yet the trajectory is produced by learned conditioning, not by being steered at sampling time. We steer the sampler of a general driving world model without knowledge-specific conditioning and retraining. 

\textbf{Rule-aware trajectory generation.} Closest to our motivation, CTG and its language-guided successor compile traffic rules into differentiable guidance for trajectory diffusion~\citep{zhong2023ctg,zhong2023ctgpp}, and Diffusion-Planner applies training-free guidance with collision, comfort, and lane energies to an ego planner~\citep{zheng2025diffusionplanner}. We differ in the model architecture. We use a frozen world model that jointly generates future driving video and planned ego trajectory. Additionally, we study whether the guided trajectory also steers the generated video through joint self-attention between the video and the ego trajectory.

\textbf{Positioning.} Our work mainly differentiates itself on the conjunction: a rectified-flow world model guided by training-free sampling-time energy of its jointly generated ego plan. The closest neighbors fall on either side: training-free energy-guided ego-plan optimization exists, but for trajectory-only models~\citep{zheng2025diffusionplanner}. World models that jointly denoise video and an ego plan exist, but inject knowledge only at training time~\citep{zhang2025epona,xia2025drivelaw}. None steers a jointly generated ego trajectory at sampling time, and we further examine whether doing so moves the generated video.

\section{Method}
\label{sec:method}

\subsection{Background: rectified-flow sampling}
We use a rectified-flow model as our backbone. For a clean sample $\xz$ and noise $\xone\sim\mathcal{N}(0,I)$, the interpolation and learned velocity field are

\begin{equation}
\begin{aligned}
\xt &= (1-t)\,\xz + \big(\sigma_{\min} + (1-\sigma_{\min})t\big)\xone,\\
\vfield_\theta(\xt,t) &\approx (1-\sigma_{\min})\xone - \xz ,
\end{aligned}
\end{equation}

and sampling integrates the ODE $\dot{\xt}=\vfield_\theta$ from $t{=}1$ to $0$. Here, $\sigma_{\min}$ is a small noise floor at the data end ($t{=}0$): flow matching models data as a narrow Gaussian of width $\sigma_{\min}$ rather than a point mass~\citep{lipman2023flowmatching}. This keeps the score and the clean-sample inversion well-defined. Pure rectified flow~\citep{liu2023rectifiedflow} is the $\sigma_{\min}{=}0$ limit, and we use $\sigma_{\min}{=}10^{-5}$. With $\sigma_{\min}\!\to\!0$ the model admits a one-step estimate of the clean sample,

\begin{equation}
\xhat(\xt,t) \;=\; \xt - t\,\vfield_\theta(\xt,t).
\label{eq:x0hat}
\end{equation}

\subsection{Training-free energy guidance on the ego trajectory}
\label{sec:guidance}
Our platform is a frozen video-diffusion world model built on a post-trained Open-Sora~2.0 MM-DiT backbone that jointly denoises a future video and a planned ego trajectory $\xz^{\mathrm{ego}}\in\R^{T\times d}$ conditioned on the observed past without agent and map information (see~\cref{fig:mmdit}, full architecture and fine-tuning details in \cref{app:training}). The energy guidance is supplied only at sampling time (see \cref{fig:overview}). We inject driving knowledge only through a differentiable energy $\mathcal{E}(\cdot)$ evaluated on the predicted clean ego trajectory. At each sampling step, we form $\xhat$ via \cref{eq:x0hat}, take an energy-descent step in clean-trajectory space, and convert it back to a velocity correction:

\begin{equation}
\begin{aligned}
\xhat' &= \xhat - \lambda(t)\,\grad_{\xhat}\mathcal{E}(\xhat),\\
\vfield' &= \frac{\xt - \xhat'}{\max(t,\,t_{\min})},
\end{aligned}
\label{eq:guide}
\end{equation}

where $\mathcal{E}$ is the driving-norm energy evaluated on the one-step estimate (see \cref{sec:energies} for an example), and $\lambda(t)$ is an asymmetric schedule that is zero near $t{=}0$ and $t{=}1$. Guidance off near $t{=}1$ lets the prior establish global structure before it is refined, while off near $t{=}0$ avoids the $1/t$ blow-up of \cref{eq:guide}.

The one-step estimate $\xhat$ is unreliable early in sampling, as the high-variance guess can sit far from the target, so the energy gradient evaluated on it can be large and poorly directed. An unclipped step launches the trajectory off-manifold. We therefore cap the per-step guidance correction at a fixed multiple of the backbone's own velocity (a scale-free clip), in line with standard training-free guidance practice~\citep{yu2023freedom,bansal2023universal}.

\subsection{Energy for driving}
\label{sec:energies}
We integrate the driving norms as differentiable energy $\mathcal{E}(\xhat)$ on the predicted clean ego trajectory. The guidance step~\eqref{eq:guide} steers the plan toward low-energy trajectories. In this paper, we use a single energy: an MSE attraction toward a counterfactual braking target in Eq.~\eqref{eq:mse}. Similar to other training-free guidance~\citep{chung2023dps,he2024mpgd}, Eq.~\eqref{eq:guide} biases the generative distribution toward low-energy samples rather than sampling exactly from an energy-tilted target, and it relies on the one-step estimate $\xhat$, which is least accurate early in sampling and under multimodality, further discussed in~\cref{sec:discussion}.

\begin{figure}[htb]
  \centering
  \includegraphics[width=\linewidth]{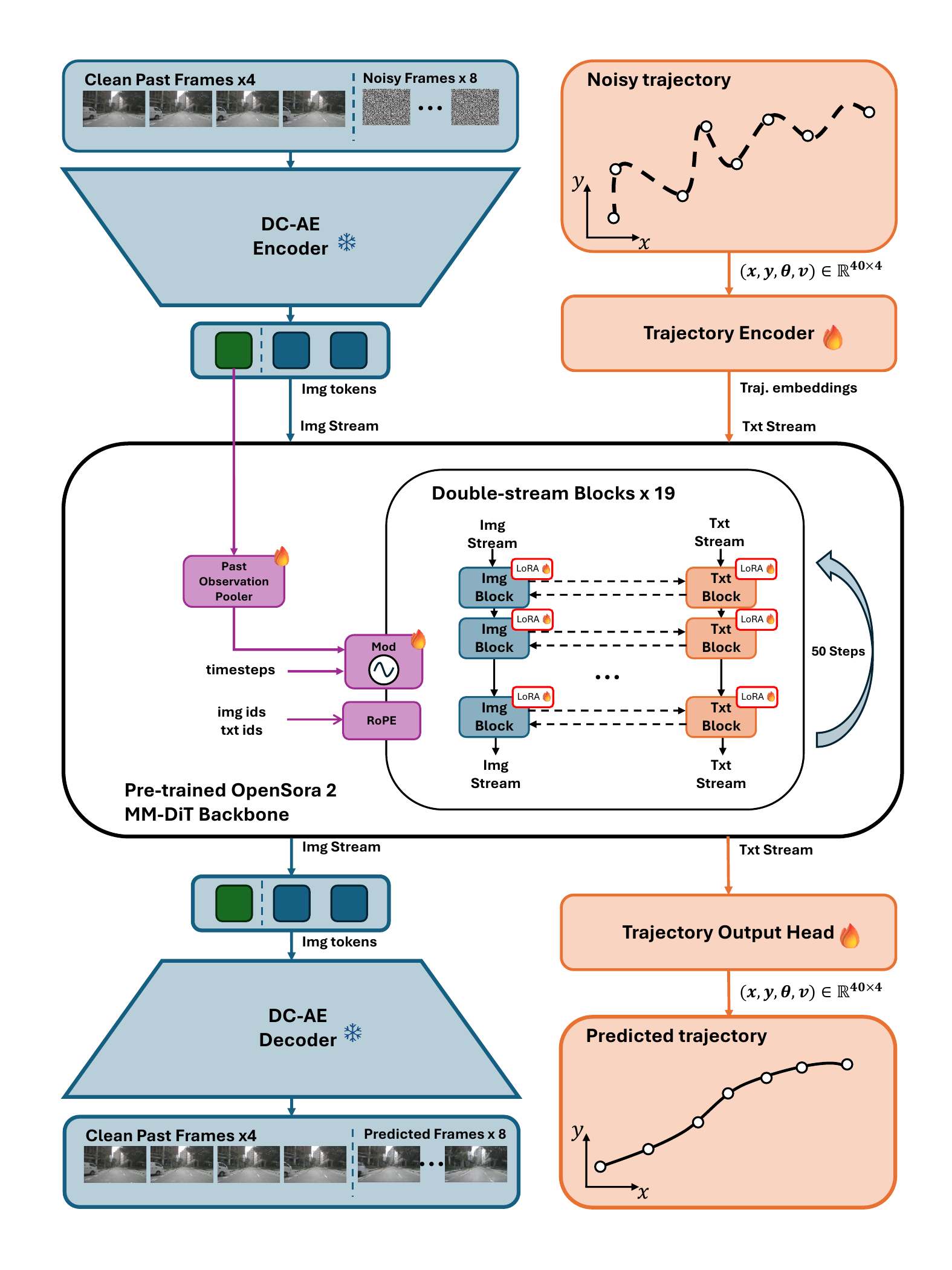}
  \caption{\textbf{The MM-DiT backbone at sampling time without energy guidance}.
  The observed past and the noised future frames are encoded by the frozen DC-AE into image tokens; the noised ego trajectory $(x,y,\theta,v)$ enters the planning (text) stream. The two streams are denoised jointly across the $19$ double-stream blocks over $50$ steps, modulated by the timestep and the past-observation pooler, with RoPE positional ids, and then decoded into the predicted future frames and the predicted trajectory. The model is post-trained unguided as a driving world model: frozen modules are marked \faSnowflake{}; modules adapted with LoRA or trained from scratch are marked \faFire{}.}
  \label{fig:mmdit}
\end{figure}

We take the notation from~\cref{sec:guidance} and~\cref{fig:mmdit}: the generated planned ego trajectory $\xz^{\mathrm{ego}} = (\mathbf{x}, \mathbf{y}, \mathbf{\theta}, \mathbf{v})\in\R^{T\times d}$ $(T=40, d=4)$

\begin{equation}
\begin{aligned}
\mathcal{E}(\xhat)=\frac{1}{|T|}\sum_{k\in T}\Big[w_{xy}\,\big\lVert (x_k,y_k)-(x_k^{\star},y_k^{\star})\big\rVert^2 \\
+w_{\theta}\,(\theta_k-\theta_k^{\star})^2 +w_{v}\,(v_k-v_k^{\star})^2\Big],
\end{aligned}
\label{eq:mse}
\end{equation}

where $(x_k,y_k,\theta_k,v_k)$ are the per-frame components of the predicted clean ego trajectory $\xhat$ (the text stream channel of \cref{fig:mmdit}), $(\cdot)^{\star}$ is the counterfactual target trajectory, which in this case simulates a slow lead vehicle in front of the ego, $T$ the valid future frames, and $w_{xy},w_{\theta},w_{v}$ per-channel weights.

\section{Experiments}
\label{sec:experiments}

\begin{figure*}[t]
  \centering
  \includegraphics[width=\linewidth]{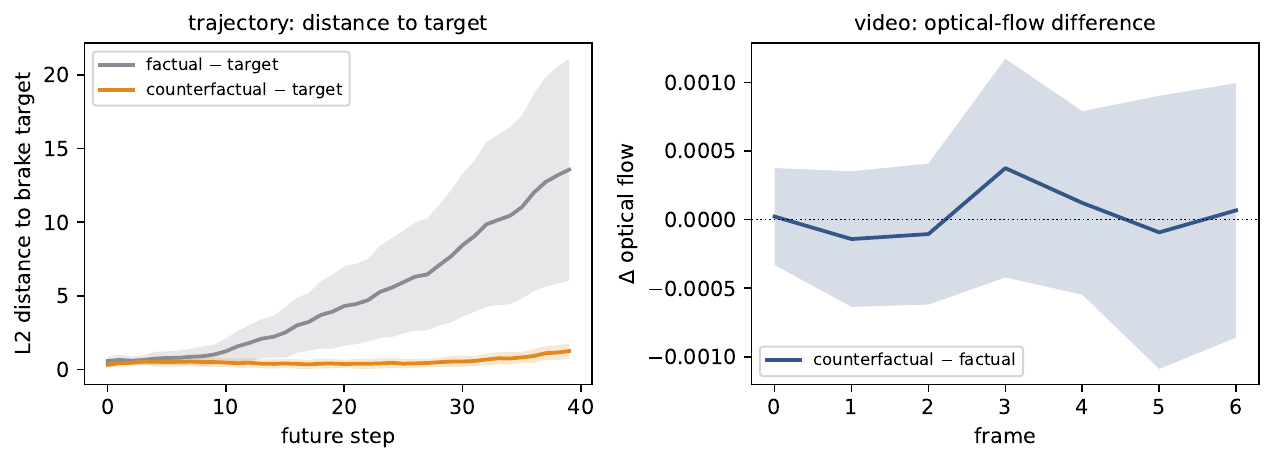}
  \caption{\textbf{The trajectory steers, the video does not.} Across $35$ sampled clips at $\lambda{=}40$ (mean $\pm$ std). For each clip, we sample the platform twice from identical noise - unguided (factual) and with a counterfactual brake energy (counterfactual) - and compare. \emph{Left:} $L_2$ distance of each ego plan to the brake target per step; the counterfactual is driven to the target while the factual diverges. \emph{Right:} per-frame optical-flow difference (counterfactual $-$ factual), which sits at zero within the across-clip spread (shaded). The steered plan does not move the pixels. Detailed results for single clips and per-frame trajectory-video comparison are evaluated and shown in~\cref{app:probe-detail}.}
  \label{fig:probe}
\end{figure*}

We apply this training-free energy guidance to the full video world model, which generates a future driving video and the ego trajectory using a counterfactual target energy that asks the planned trajectory to brake, simulating a rear-end safety driving norm (a slow lead vehicle ahead of the ego). The guidance steers the ego trajectory effectively: across $35$ sampled clips, the guided trajectory ends a mean $1.3$\,m from the brake target while the unguided trajectory ends $13.6$\,m away of it (\cref{fig:probe}, left), showing that a single sampling-time energy steers the trajectory of the full video world model. Detailed results for single clips and per-frame trajectory-video comparison are evaluated and shown in~\cref{app:probe-detail} (see~\cref{fig:probe-bev-a},~\cref{fig:probe-bev-b}, and~\cref{fig:probe-aligned}).

The open question is whether the video follows. Since the video and the trajectory are denoised jointly, the video should follow the steered trajectory through the joint self-attention mechanism from the backbone. However, we discovered that the video does not follow: at the same operating point, the per-frame optical flow of the guided video is statistically indistinguishable from the unguided one (a ${\lesssim}2\%$ change, within the across-clip spread; \cref{fig:probe}, right), even though the trajectory has braked. We suspect that the problem comes from the fact that the $38$ single-stream blocks are currently bypassed in our modified model, where the video and trajectory streams would be processed jointly (see \cref{app:backbone}). Our current proposed solution is to re-route the video and trajectory streams back to the $38$ single-stream blocks to align the video and trajectory streams. 

\section{Discussion, Limitations, and Ongoing Work}
\label{sec:discussion}

\textbf{Limitations of the mechanism.} Our guidance in Eq.~\eqref{eq:guide} biases the prior toward low-energy samples, which inherits the known limitations of training-free guidance: the one-step estimate $\xhat$ is least accurate early in sampling and under multi-modality \cite{chung2023dps, he2024mpgd}. Furthermore, ODE sampling lacks the stochastic self-correction that helps score-SDE samplers return to the data manifold \cite{song2021scoresde}. The asymmetric schedule in \cref{sec:guidance} mitigates but does not eliminate these known limitations completely. We also stabilize the guidance by clipping the per-step correction early in the stage, as the energy gradient is unreliable and $\xhat$ can be large (see~\cref{app:guidance-vel}).

\textbf{Ongoing work: making the video follow.} We established the formulation and showed that sampling-time energy steers the ego trajectory of the world model. The current problem comes from the cross-stream control: the steered planned trajectory does not yet move the video, because the ego interacts with the video only in the double-stream blocks through the joint self-attention. We plan to fix the cross-stream problem by routing both streams through the single-stream blocks from the Open-Sora~2.0 backbone, where the two streams are processed jointly. This could, in turn, re-anchor the ego trajectory to overlap with the generated video. Furthermore, we plan to run further evaluation on the generation realism (e.g., FVD \cite{unterthiner2018fvd}) and rule compliance.

\textbf{Ongoing work: composing multiple driving norms.} A single energy already steers the plan; our final goal is to express the full body of driving knowledge, including collision avoidance, lane keeping, speed limits, and comfort, as a library of differentiable energies injected jointly at sampling time. The difficulty is that such norms routinely conflict: a comfort term that smooths a trajectory may oppose a safety term that demands hard braking. A plain weighted sum of their gradients lets a lower-priority objective override a higher one whenever its magnitude happens to dominate. We therefore plan to impose a priority hierarchy (safety $>$ traffic rules $>$ comfort) at the level of the guidance gradient rather than through hand-tuned weights, which means that lower-priority gradients are projected onto the subspace that does not oppose the active higher-priority constraints at each step. Per-level magnitudes are normalized so that no choice of weights can let, e.g., comfort defeat a safety rule. This would reconcile heterogeneous, possibly conflicting driving rules into a single composed energy field with guaranteed precedence, while preserving the training-free property that a deployed model can add or re-prioritize a rule purely at sampling time.

\textbf{Conclusion.} We show that driving-norm compliance need not be trained in: a frozen rectified-flow world model can have its planned ego trajectory steered at sampling time by differentiable energies. We demonstrate this by steering the trajectory to a counterfactual braking target, which the model is not trained to condition on. The open question fixates on the video stream: under the current architecture, the video does not yet follow the steered trajectory, which we attribute to the cross-stream coupling, and we propose that deep joint denoising of the two streams is the next step toward end-to-end controllable rollouts.

\clearpage

\bibliographystyle{ieeetr}
\bibliography{references}

\clearpage

\appendix  
This appendix derives the energy guidance for rectified flow used in~\cref{sec:method} (§~\cref{app:rf-guidance}), documents the model architecture and fine-tuning details (§~\cref{app:training}), and illustrates the energy guidance on video and ego-trajectory generation (§~\cref{app:probe-detail}).

\subsection{Energy Guidance for Rectified Flow}
\label{app:rf-guidance}

This section first establishes the equivalency between the velocity field in rectified flow and the marginal score from the energy score in ~\cref{app:score-vel-equivalency}, and discusses the implementation and the overhead of the energy guidance in~\cref{app:guidance-vel} and \cref{app:cost}.

Our backbone is a rectified-flow model with the interpolant and conditional velocity target

\begin{equation}
\begin{aligned}
\xt = \alpha_t\,\xz + \sigma_t\,\xone,\quad \xone\sim\mathcal N(0,I),\\
\alpha_t = 1-t,\quad
\sigma_t = \sigma_{\min} + (1-\sigma_{\min})\,t,\quad
\end{aligned}
\label{eq:app-interp}
\end{equation}

with $\vfield_\theta(\xt,t)\approx\mathbb E[\dot{\xt}\mid\xt]=(1-\sigma_{\min})\xone-\xz$. The floor $\sigma_{\min}$ is the minimum noise at the data end~\citep{lipman2023flowmatching}. Open-Sora~2.0 uses $\sigma_{\min}{=}10^{-5}$ with the Movie Gen parameterization~\citep{peng2025opensora2}. We take the same formulation for the interpolant, the velocity target, this $\sigma_{\min}$, and the timestep schedule at fine-tuning and sampling, so the schedule and $\sigma_{\min}$ match those the backbone was trained with.

\subsubsection{Score and velocity equivalency}
\label{app:score-vel-equivalency}

For the Gaussian path, the velocity field and the marginal score $s_t(\xt)=\grad_{\xt}\log p_t(\xt)$ are affinely equivalent~\citep{feng2025guidanceflowmatching,albergo2023stochastic}:

\begin{equation}
\begin{aligned}
\vfield_\theta(\xt,t) = -\frac{\xt + \sigma_t\, s_t(\xt)}{\alpha_t}, \\
\qquad
s_t(\xt) = -\,\frac{\alpha_t\,\vfield_\theta(\xt,t) + \xt}{\sigma_t}.
\end{aligned}
\label{eq:app-bridge}
\end{equation}

The transformation is affine and invertible at every $t\in(0,1)$, so any score-based guidance transfers to the velocity field exactly, without score network or retraining. The one-step clean-sample estimate is the rectified-flow analogue of Tweedie's posterior mean, $\xhat=\xt-t\,\vfield_\theta$ as $\sigma_{\min}\!\to\!0$~(same as~\cref{eq:x0hat}).

\subsubsection{Guided velocity.}
\label{app:guidance-vel}
Score-based energy guidance encodes external knowledge as a likelihood $p_t(y\mid\xt)$. With the standard training-free approximation ($\grad_{\xt}\log p_t(y\mid\xt)\approx-\grad_{\xt}\mathcal E(\xhat)$) and the affine transformation~\cref{eq:app-bridge}, together with the constant Jacobian $\partial\xhat/\partial\xt=(1-\sigma_{\min})I$, the guided velocity is a single additive correction in clean-sample space,

\begin{equation}
\begin{aligned}
\tilde{\vfield}(\xt,t) = \vfield_\theta(\xt,t) + w(t)\,\grad_{\xhat}\mathcal E(\xhat),\qquad \\
w(t) = \frac{(1-\sigma_{\min})\,\sigma_t}{\alpha_t},
\end{aligned}
\label{eq:app-guided}
\end{equation}

followed by the ordinary Euler step. We replace the weight $w(t)$ with the schedule $\lambda(t)$ in \cref{eq:guide}, which is capped near $t{=}1$ and zeroed near $t{=}0$, as weight $w(t)$ grows without bound there. We allow $\lambda(t)$ to be a tunable hyperparameter during sampling in the implementation. 

\subsubsection{Energy guidance overhead during sampling}
\label{app:cost}

The correction needs only $\grad_{\xhat}\mathcal E$, which we obtain with ordinary automatic differentiation (PyTorch autograd) on the small closed-form energy of the $(x,y,\theta,v)$ ego trajectory. Because the energy is evaluated on $\xhat$ with the predicted velocity detached, the $11\mathrm{B}$-parameter backbone never enters the \texttt{autograd} graph. Moreover, because the guided quantity is the trajectory rather than the video latents, there is no VAE backpropagation and no pixel-space manifold drift. The cost is one gradient of a small analytic function per step, therefore negligible beside the forward pass.

\subsection{Model Architecture and Fine-tuning Details}
\label{app:training}

This section documents how the video-diffusion driving world model in~\cref{sec:method} is obtained. All numbers are for the energy-guided platform evaluated in the main text; the flow-matching convention and timestep schedule are those of \cref{app:rf-guidance}. The backbone's sampling-time computation graph is shown in \cref{fig:mmdit}.

\subsubsection{Backbone and Architectural Modifications}
\label{app:backbone}
The backbone is the Open-Sora~2.0 MM-DiT (DC-AE variant)~\citep{peng2025opensora2}: $19$ double-stream blocks and $38$ single-stream blocks, hidden size $3072$, $24$ heads, with separate $q/k/v$ projections. The image stream predicts the future videos. We repurpose the text stream into ego trajectory prediction. Each stream gets its own timestep embedding. The video reuses the pretrained timestep MLP, while the ego timestep MLP is newly added and zero-initialized, so at initialization (tied timesteps) the model reproduces the pretrained behaviour exactly. A past-observation pooler maps the clean past latent to the conditioning channel. The  single-stream blocks are \textit{bypassed} due to architectural modifications for the ego trajectory prediction, because we need to output both video and trajectory prediction at the same time, and the $19$ double-stream blocks provide such interfaces.

\subsubsection{Fine-tuning details}
The entire backbone is frozen. We attach LoRA adapters~\citep{hu2022lora} (rank $16$, $\alpha=32$) to the modulation, attention, and MLP projections of both streams (\texttt{img\_\{mod,attn,mlp\}}, \texttt{txt\_\{mod,attn,mlp\}}, $266$ linear layers, $46.7$M parameters trained at learning rate $1\!\times\!10^{-4}$). The newly introduced modules (ego encoder, ego output head, ego timestep MLP, past-obs routing, and pooler) are initialized with zeros and trained from scratch at $3\!\times\!10^{-4}$. In total $88.1$M parameters are trainable out of the $\sim\!11$B backbone.

\subsubsection{Objective and optimization}
We minimize a joint rectified-flow loss $\mathcal{L} = \mathcal{L}_{\text{video}} + \lambda_{ego} \mathcal{L}_{\text{ego}}$ ($\lambda_{ego}=1$), each an MSE on the predicted velocity (\cref{app:rf-guidance}). The ego term is masked to the valid future frames. Optimization uses AdamW ($\beta=(0.9,0.98)$, weight decay $0$, $\epsilon=10^{-8}$) with gradient clipping at $1.0$ and \texttt{bf16} autocast. To fit the backbone on a single NVIDIA HGX H100 94GB GPU we keep the frozen weights in \texttt{bf16} and apply gradient activation checkpointing to every transformer block. Together, these allow the batch-$8$ training to fit on the single card.

\subsubsection{Classifier-free guidance}
The platform is also trained for classifier-free guidance \cite{ho2022classifierfree} over its past conditioning, which sharpens the generated video. With probability $0.15$ per sample we drop the observed past by replacing the pooled past-observation vector with a learned null embedding and zeroing the in-context past latent frames together, so the model additionally learns the unconditional (marginal) future distribution. At sampling we extrapolate $\vfield = \vfield_{\varnothing} + w\,(\vfield_{\mathrm{c}}-\vfield_{\varnothing})$ on both streams with guidance scale $w$ (we use $w\!\in\![3,5]$). This is orthogonal to the energy guidance of \cref{sec:method}, which acts only on the ego stream and needs no retraining.

\subsubsection{Data and autoencoder}
We train on nuScenes \texttt{v1.0-trainval} \cite{caesar2020nuscenes}, front camera, at $256\!\times\!448$, using $12$-frame clips at the native sample-data cadence (stride $12$, non-overlapping). The frozen DC-AE (\texttt{dc-ae-f32t4c128}, scaling factor $0.493$, $4\times$ temporal / $32\times$ spatial compression) encodes each clip to a $(3,8,14)$ latent, where the first latent frame is the clean past conditioning and the remaining two are the denoised future. The ego target is the $40$-step future trajectory.

\subsubsection{Schedule and checkpoint}
Timesteps are sampled logit-normal (in compliance with the Open-Sora~2.0 backbone); the batch size is $8$ per GPU. The checkpoint used here is trained for $\sim\!22$k steps on a single GPU, with validation on a held-out tail of $64$ clips and best-on-validation checkpointing. The platform is trained once. Every result in the main text reuses these frozen weights and adds knowledge only through sampling-time energies.

\subsubsection{Code availability}
We release the code required to reproduce the results from the released checkpoint: the training-free energy-guidance sampler, the differentiable energy library and the score-velocity bridge, the counterfactual-brake probe, and the analysis and figure scripts at \url{https://timsu-98.github.io/energy_guidance_for_driving_world_models/}. We also release the trained model weights with the LoRA adapters and the additionally introduced modules ($88$M trainable parameters). Reconstituting the full backbone also requires two checkpoints that we use and do not redistribute: the $\sim\!11$B Open-Sora~2.0 base MM-DiT~\citep{peng2025opensora2} and its DC-AE autoencoder (\texttt{dc-ae-f32t4c128}), both obtained from their original public releases.

\subsection{Counterfactual Probe: Per-Clip Views}
\label{app:probe-detail}

\begin{figure*}[p]
    \centering
    \includegraphics[width=\textwidth]{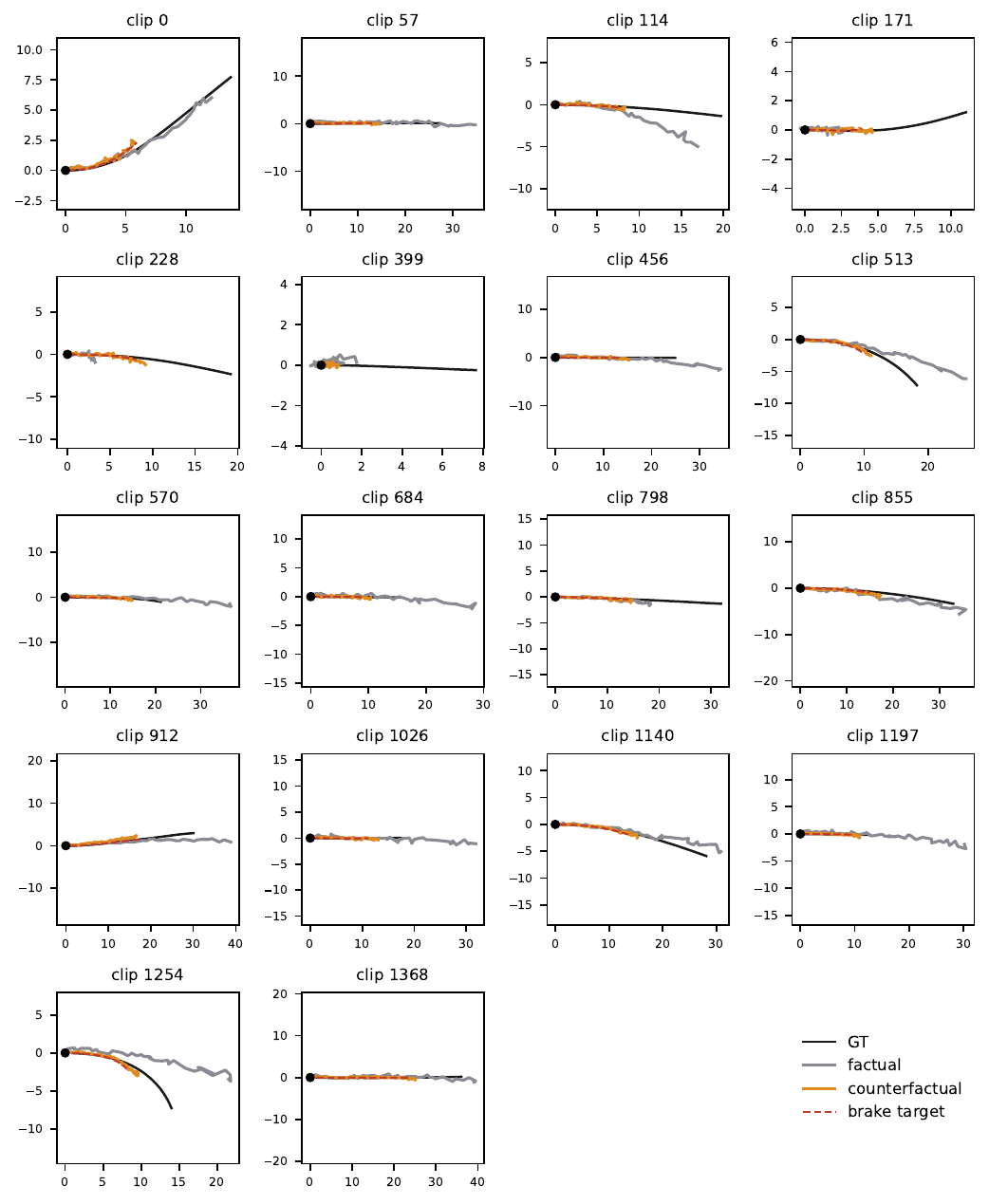}
    \caption{\textbf{Per-clip BEV trajectories (1/2).} All $35$ usable sampled clips at
    $\lambda{=}40$. GT (black), unguided/factual (grey), counterfactual (orange), brake
    target (red dashed). The counterfactual plan tracks the braked target across clips.}
    \label{fig:probe-bev-a}
\end{figure*}

\begin{figure*}[p]
    \centering
    \includegraphics[width=\textwidth]{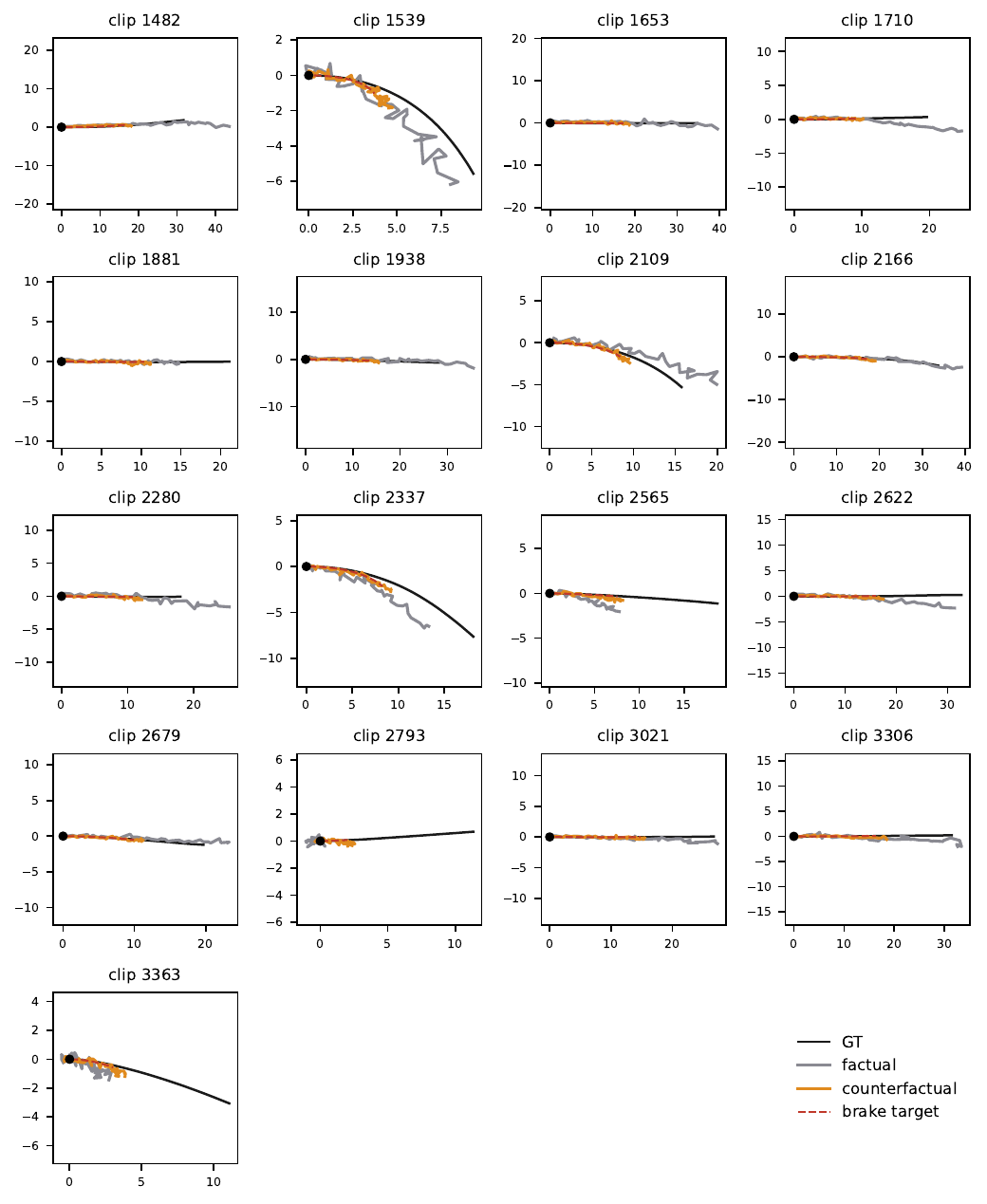}
    \caption{\textbf{Per-clip BEV trajectories (2/2).} Continuation of \cref{fig:probe-bev-a}.}
    \label{fig:probe-bev-b}
\end{figure*}

\begin{figure*}[t]
  \centering
  \includegraphics[width=0.85\textwidth]{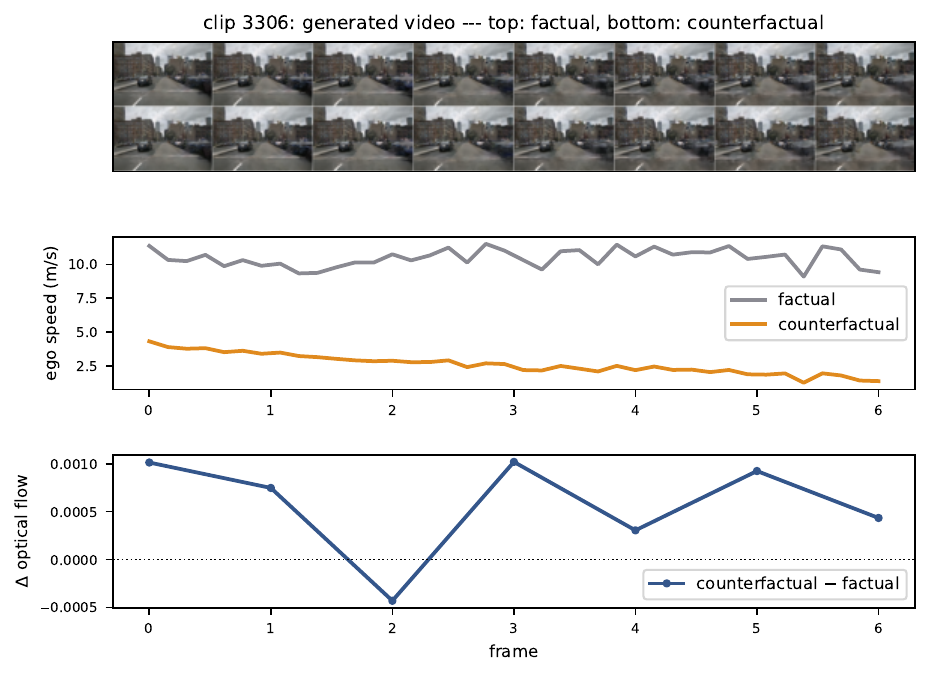}
  \caption{\textbf{One clip, aligned on frames.} \emph{Top:} ego speed - the counterfactual brakes hard while the factual holds speed. \emph{Bottom:} per-frame optical-flow difference (counterfactual $-$ factual), at zero: the trajectory steers but the video does not. (Plan steps are mapped onto the video-frame axis for alignment.)}
  \label{fig:probe-aligned}
\end{figure*}

\Cref{fig:probe} aggregates the cross-stream probe. Here, we show its per-clip structure. \Cref{fig:probe-bev-a} and \cref{fig:probe-bev-b} overlay the bird's-eye-view trajectories: the counterfactual plan (orange) tracks the braked target while the unguided plan (grey) runs long, on essentially every clip. \Cref{fig:probe-aligned} aligns one clip's ego speed and the per-frame optical-flow difference on a common frame axis: the plan brakes hard while the optical-flow difference stays around zero, which indicates the steered planned trajectory does not reach the video stream.


\end{document}